\documentclass[10pt,twocolumn,letterpaper]{article}

\usepackage{cvpr}
\usepackage{times}
\usepackage{epsfig}
\usepackage{graphicx}
\usepackage{amsmath}
\usepackage{amssymb}
\usepackage[]{algorithm2e}

\newcommand{\devis}[2]{\textcolor [RGB]{150,50,50}{#2}}

\usepackage[pagebackref=true,breaklinks=true,letterpaper=true,colorlinks,bookmarks=false]{hyperref}

\cvprfinalcopy 

\def\cvprPaperID{1883} 

\ifcvprfinal\fi
\begin{document}

\title{Learning deep structured active contours end-to-end}

\author{\normalsize Diego Marcos, Devis Tuia, Benjamin Kellenberger\\
\normalsize University of Wageningen, Netherlands\\
{\tt\small name.surname@wur.nl}
\and
\normalsize Lisa Zhang, Min Bai, Renjie Liao, Raquel Urtasun\\
\normalsize University of Toronto, Canada\\
{\tt\small \{lczhang,mbai,rjliao,urtasun\}@cs.toronto.edu}
}

\maketitle

\begin{abstract}
The world is covered with millions of buildings, and precisely knowing each instance's position and extents is vital to a multitude of applications. Recently, automated building footprint segmentation models have shown superior detection accuracy thanks to the usage of Convolutional Neural Networks (CNN). However, even the latest evolutions struggle to precisely delineating borders, which often leads to geometric distortions and inadvertent fusion of adjacent building instances.
We propose to overcome this issue by exploiting the distinct geometric properties of buildings. To this end, we present Deep Structured Active Contours (DSAC), a novel framework that integrates priors and constraints into the segmentation process, such as continuous boundaries, smooth edges, and sharp corners. To do so, DSAC employs Active Contour Models (ACM), a family of constraint- and prior-based polygonal models. We learn ACM parameterizations per instance using a CNN, and show how to incorporate all components in a structured output model, making DSAC trainable end-to-end. We evaluate DSAC on three challenging building instance segmentation datasets, where it compares favorably against state-of-the-art. Code will be made available on \url{https://github.com/dmarcosg/DSAC}.

\end{abstract}

\section{Introduction}
Accurate footprints of individual buildings are of paramount importance for a wide range of applications, such as census studies~\cite{xie2015population}, disaster response after earthquakes~\cite{sahar2010using} and developmental assistances like malaria control~\cite{franke2015earth}. Automating large-scale building footprint segmentation has thus been an active research field, and the emergence of high-capacity models like fully convolutional networks (FCNs)~\cite{gupta2014learning}, together with vast training data~\cite{wang2016torontocity}, has led to promising improvements in this field.

\begin{figure}[t]
\begin{tabular}{ccc}
\includegraphics[width=0.28\linewidth]{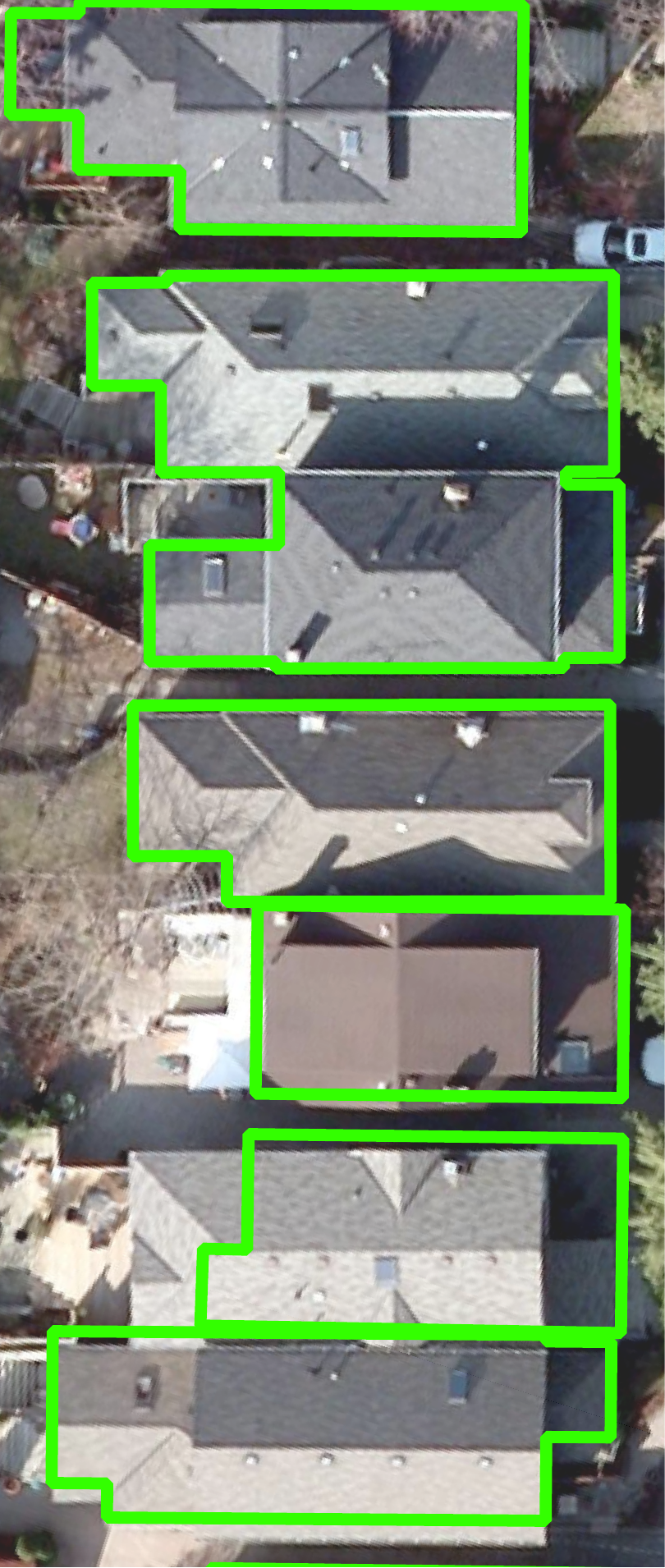} &
\includegraphics[width=0.28\linewidth]{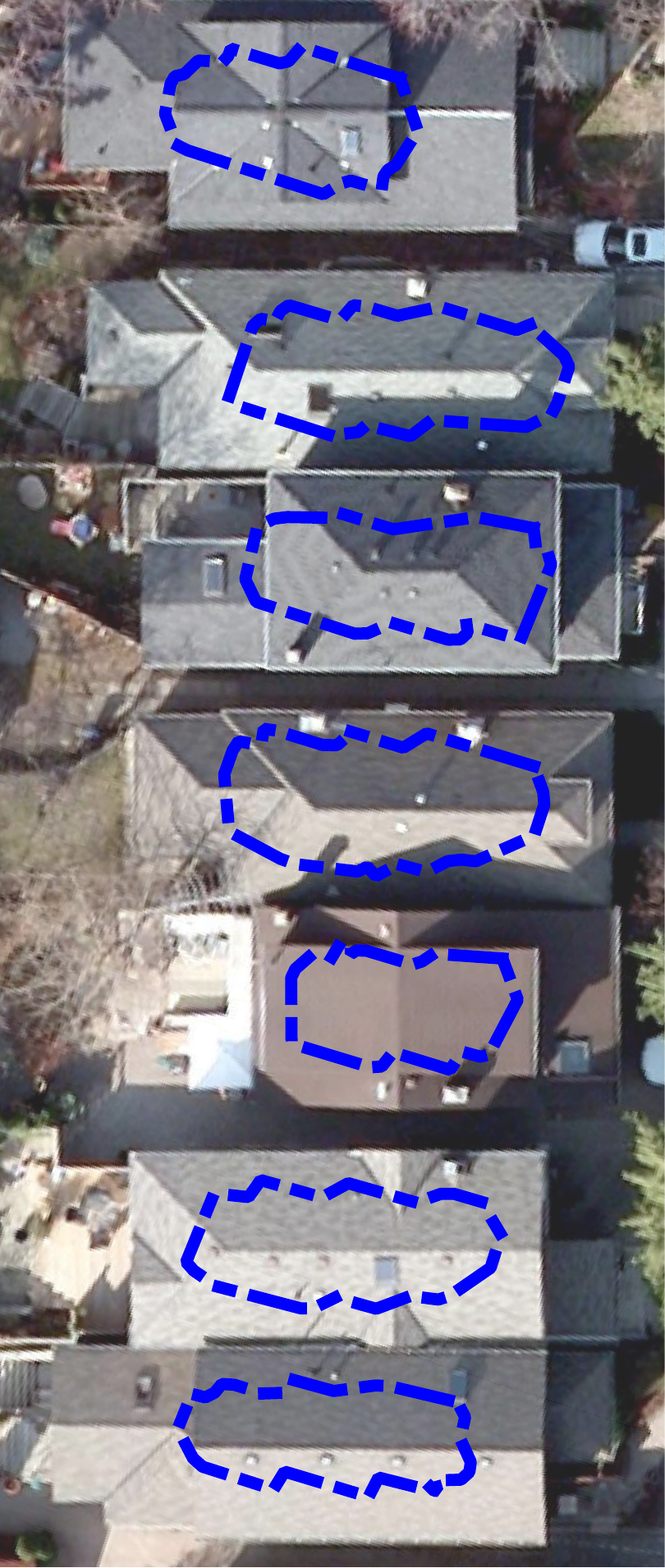} &
\includegraphics[width=0.28\linewidth]{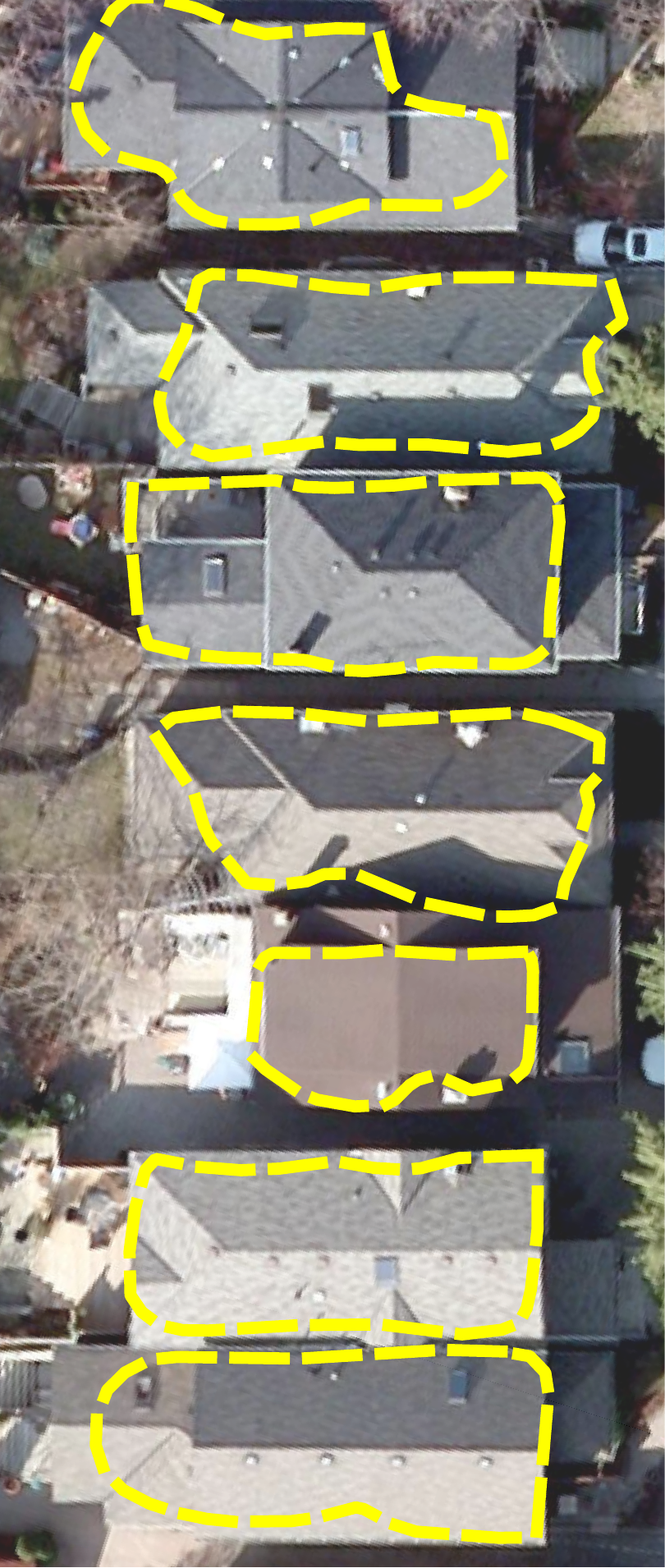}\\
GT & Init. & Result
\end{tabular}
\caption{DSAC uses a CNN to predict the energy function used by an Active Contour Model (ACM) to modify an initial instance polygon using learned geometric priors. Left: image from the TorontoCity validation dataset with ground truth polygons, center: initial polygons provided by~\cite{Bai17}, right: results of DSAC.}
\end{figure}

Most studies address semantic segmentation of buildings, which consists of inferring a class label (e.g. ``building'') densely for each pixel over the overhead image of interest~\cite{kaiser2017learning,maggiori2017convolutional,montoya2015semantic,volpi2017dense}. While this approach may provide global statistics such as building area coverage estimation, it comes short at yielding estimations at the instance level. In computer vision, this problem is known as \emph{instance segmentation}, where models provide a segmentation mask on a per-object instance basis. Solving this task is far more challenging than semantic segmentation, since the model has to understand whether any two building pixels belong to the same building or not. Precise delineation of object borders, with sharp corners and straight walls in the case of buildings, is a task that CNNs generally perform poorly at~\cite{dai2016r}: as a result, building segmentations from CNNs commonly have a high detection rate, but fail in terms of spatial coverage and geometric correctness.

Active Contour Models (ACM~\cite{kass1988snakes}), also called \emph{snakes}, may be considered to address this issue. ACMs augment bottom-up boundary detectors with high-level geometric constraints and priors. They work by constraining the possible outputs to a family of curves (e.g. closed polygons with a fixed number of vertices), and optimizing them by means of energy minimization based on both the image features and a set of shape priors such as boundary continuity and smoothness. Additional terms have been proposed, among which the balloon term~\cite{cohen1991active} is of particular interest: it mimics the inflation of a balloon by continuously pushing the snakes' vertices outwards, thus preventing it to collapse to a single point. 
By expressing object detection as a polygon fitting problem with prior knowledge, ACMs have the potential of approaching object edges precisely and without the need for additional post-processing. However, the original formulation lacked flexibility, since it relied on low-level image features and a global parameterization of priors, when a more useful approach would be to penalize strongly the curvature in the regions of the boundary known to be straight or smooth and reduce the penalization in the regions that are more likely to form a corner. Moreover, the balloon term has so far only been included as a post-energy global minimization force and does not take part in the energy minimization defining the snake.

In this paper, we propose to combine the expressiveness of deep CNNs with the versatility of ACMs in a unified framework, which we term Deep Structured Active Contours (DSAC). In essence, we employ a CNN to learn the energy function  that would allow an ACM to generate polygons close to a set of ground truth instances. To do so, DSAC leverages the original ACM formulation by learning high-level features and prior parameterizations, including the balloon term, in one model and \emph{on a local basis, i.e.} penalizing each term differently at each image location. We cast the optimization of the ACM as a structured prediction problem and find optimal features and parameters using a Structured Support Vector Machine (SSVM~\cite{Alt07,Tso05}) loss. As a consequence, DSAC is trainable end-to-end and able to learn and adapt to a particular family of object instances. We test DSAC in three building instance segmentation datasets, where it outperforms state-of-the-art models.

\paragraph{Contributions} This work's contributions are as follows:
\begin{itemize}
\item We formulate the learning of the energy function of an ACM as a structured prediction problem;
\item We include the balloon term of the ACM into the energy formulation;
\item We propose an end-to-end framework to learn the guiding features and local priors with a CNN.
\end{itemize}


\section{Related work}
\label{sec:relWork}

\paragraph{Building footprint extraction}

Most current automated approaches make use of 3D information extracted from ground or aerial LIDAR~\cite{wang2006bayesian}, or employ humans in the loop~\cite{brooks2015semi}. The use of a polygonal shape prior has been shown to substantially improve the results~\cite{sun2014free} of systems based on color imagery and low level features. 
Recent efforts employ deep CNNs for semantic segmentation and allowed a great leap towards full automation of building segmentation~\cite{kaiser2017learning}. Works considering building instance segmentation are scarcer and the task has been recently defined as far-from-being solved~\cite{wang2016torontocity}, despite the interest shown by the participation to numerous contests aiming at automatic vectorization of building footprints from overhead imagery: SpaceNet\footnote{\hyperref[]{https://wwwtc.wpengine.com/spacenet}}, DSTL\footnote{\hyperref[]{https://www.kaggle.com/c/dstl-satellite-imagery-feature-detection}} or  OpenAI Challenge\footnote{\hyperref[]{https://werobotics.org/blog/2018/01/10/open-ai-challenge/}}. Our proposed DSAC aims at making high-level geometric information available to CNN based methods as a step towards bridging this gap.

\paragraph{Instance segmentation in Computer Vision}

Since instance segmentation combines object detection and dense segmentation, many proposed pipelines attempt at fusing both tasks in either separate or end-to-end trainable models. For example,~\cite{dai2016instance} employ a multi-task CNN to detect candidate objects and infer segmentation masks and class labels per detection. 
\cite{fathi2017semantic} train a CNN on pairs of locations and predicts the likelihood for the pair to belong to the same object. \cite{romera2016recurrent} apply an attention-based RNN sequentially on deep image features to trace object instances in propagation order. \cite{Bai17} refine an existing semantic segmentation map by predicting a distance transform to the nearest boundary. High level relationships are accounted for in~\cite{royer2016convexity,zhang2016instance} by means of an instance MRF applied to the CNN's output.

All these methods employ pixel-wise CNNs and are thus not apt to integrating output shape priors directly, as polygonal output models would be. Only a few works deal with CNNs that explicitly produce a polygonal output. In~\cite{Cas17}, a recursive neural network is used to generate a segmentation polygon node by node, while in~\cite{rupprecht2016deep} a CNN predicts the direction of the nearest object boundary for each node in a polygon and  uses it as a data term in an ACM. However, the first model is tailored towards a different problem (interactive segmentation and correction) and does not allow the inclusion of strong priors, and the second decouples the CNN training from ACM inference, thus lacking the end-to-end training capabilities of the proposed DSAC.

\paragraph{Active contours}
The first ACMs were introduced by Kass \emph{et al.} in 1988 under the name of snakes~\cite{kass1988snakes}. 
Variants of this original try to overcome some of its limitations, such as the need for precise initializations, or the dependence on user interaction. In \cite{gunn1997robust} the authors propose to use two coupled snakes that better capture the information in the image. The above mentioned balloon force was introduced by \cite{cohen1991active}.

Although some modifications~\cite{kichenassamy1995gradient} have been proposed to improve the data term of the original paper, they rely on simple assumptions about the appearance of the objects and on global parameters for weighting the different terms in the energy function. The proposed DSAC leverages the original formulation by including local prior information, i.e. values weighting the snakes' energy function terms on a per-pixel basis, and learns them using a CNN. Although this work focuses on curvature priors useful for segmenting objects of polygonal shape, other priors can be enforced with ACMs, such as convexity for biomedical imaging~\cite{royer2016convexity}.


\begin{figure}[t]
\includegraphics[width=\linewidth]{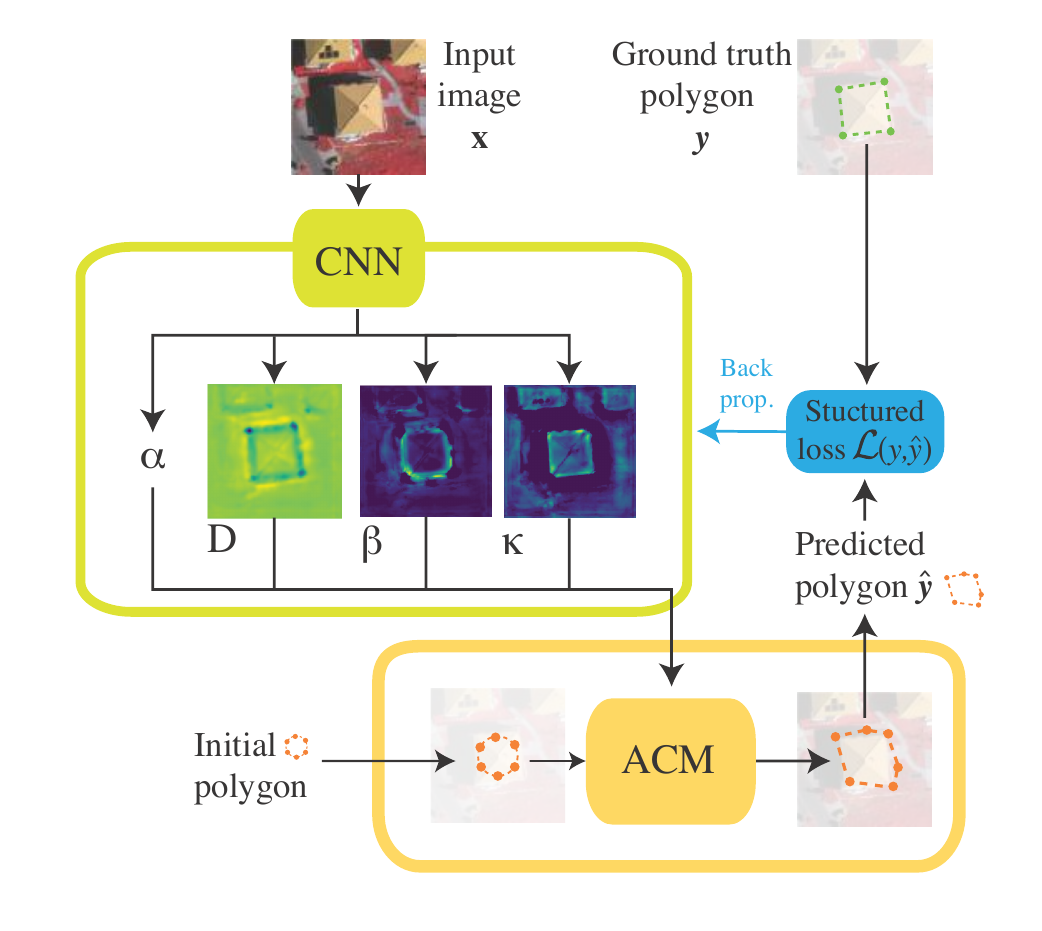}
\caption{DSAC idea. The CNN predicts the values of the energy terms to be used by the active contour model (ACM): a global $\alpha$ for the length penalization and maps for local $D$, the data term, $\beta$, the curvature penalization and $\kappa$, the balloon term. After ACM inference, a structured loss is computed and given to the CNN, whose parameters can then be updated using backpropagation. \label{fig:flow}}
\end{figure}


\paragraph{Structured learning with CNNs} Structured prediction~\cite{taskar2005learning} allows to model dependencies between multiple output variables and hence offers an elegant way to incorporate prior rule sets on output configurations. End-to-end trainable structured models exceed traditional two-step solutions by enriching the learning signal with relations at the output level. Although these models have been applied to a variety of problems~\cite{belanger2016structured,chen2015learning,schwing2015fully}, we are not aware of any work dealing with instance level segmentation.

We use a structured loss as a learning signal to a CNN such that it learns to coordinate the different ACM energy terms, which are heavily interdependent.

\section{Method}
\label{sec:method}

We present the details of a modified ACM inference algorithm with image-dependent and local penalization terms as well as the structured loss that is used to train a CNN to generate these penalization maps. A diagram of the proposed method is shown in Fig.~\ref{fig:flow}.
The proposed training algorithm proceeds as exposed in Algorithm~\ref{algo}.

\begin{algorithm}[h]
\KwData{
$\mathcal{X}, \mathcal{Y}$: image/polygon pairs in the training set.\\
$\mathcal{Y}^0$: corresponding polygon initializations.}
\For{$\mathbf{x}_i,\mathbf{y}_i \in \mathcal{X}, \mathcal{Y}$}{
CNN inference: $D$, $\alpha$, $\beta$, $\kappa \leftarrow CNN_\omega(\mathbf{x}_i)$\\
ACM inference: $\hat{\mathbf{y}}_i \leftarrow ACM(D,\alpha,\beta,\kappa,\mathbf{y}^0_i)$\\
$\frac{\partial \mathcal{L}}{\partial D}$,
$\frac{\partial \mathcal{L}}{\partial \alpha}$,
$\frac{\partial \mathcal{L}}{\partial \beta}$,
$\frac{\partial \mathcal{L}}{\partial \kappa} \leftarrow$ 
$\hat{\mathbf{y}}^i,\mathbf{y}_i$ and Eqs.~\ref{eq:grad_D}-\ref{eq:grad_K}\\
Compute $\frac{\partial \mathcal{L}}{\partial \omega}$ using backpropagation\\
Update CNN: $\omega \leftarrow \omega - \eta \frac{\partial \mathcal{L}}{\partial \omega}$
}
\label{algo}
\caption{The DSAC training algorithm. At every iteration, the CNN forward pass is followed by ACM inference, which yields a contour that is used to compute the structured loss.}
\end{algorithm}

Note that i) DSAC does not depend on any particular ACM inference algorithm, and ii) the chosen ACM algorithm does not need to be differentiable.


\subsection{Locally penalized active contours}
\label{sec:acm}

An active contour~\cite{kass1988snakes} can be represented as a polygon $\mathbf{y}=(\mathbf{u},\mathbf{v})$ with $L$ nodes $\mathbf{y}_s = (u_s,v_s) \in \mathbb{R}^2$, with $s\in {1\dots L}$, where each $s$ represents one of the nodes of the discretized contour. The polygon $\mathbf{y}$ is then deformed such that the following energy function is minimized:

\begin{multline}
	E(\mathbf{y},\mathbf{x}) = \sum_{s=1}^{L}\bigg[ D\big(\mathbf{x},(\mathbf{y}_s)\big)   
	+  
	\alpha\big(\mathbf{x},(\mathbf{y}_s)\big) \Big|\frac{\partial\mathbf{y}}{\partial s}\Big|^2  
	+ \\
	\beta\big(\mathbf{x},(\mathbf{y}_s)\big) \Big|\frac{\partial^2\mathbf{y}}{\partial s^2}\Big|^2 \bigg] +
	\sum_{u,v\in\Omega(\mathbf{y})}\kappa(\mathbf{x},(u,v)),
\label{eq:snakes}
\end{multline}

where $D\big(\mathbf{x}\big)\in \mathbb{R}^{U\times V}$ is the data term, depending on input image, of size $U\times V$, $\mathbf{x}\in \mathbb{R}^{U\times V\times d}$, $\alpha\big(\mathbf{x}\big), \beta\big(\mathbf{x}\big)\in \mathbb{R}^{U\times V}$ are the terms encouraging short and smooth polygons respectively, $\kappa(\mathbf{x})$ is the balloon term and $\Omega(\mathbf{y})$ is the region enclosed by $\mathbf{y}$. The notation $D\big(\mathbf{x},(\mathbf{y}_s)\big)$ means the value in $D\big(\mathbf{x}\big)$ indexed by the position $\mathbf{y}_s = (u_s,v_s)$. 

Due to their local nature, $D$,$\beta$ and $\kappa$ are $U\times V$ maps in our experiments while $\alpha$ is treated as a single scalar.

\subsubsection{Data term}
\label{sec:D}

This term identifies areas of the image where the nodes of the polygon should lie. In the literature, $D\big(\mathbf{x}\big)$ is usually some predefined function on the image, typically related to the image gradients. $D(\mathbf{x})$ should learn to provide relatively low values along the boundary of the object of interest and high values elsewhere. During ACM inference, the direction of steepest descent $-\nabla D(\mathbf{x}) =-\big[\frac{\partial D(\mathbf{x})}{\partial u},\frac{\partial D(\mathbf{x})}{\partial v}]$ is used as the data force term, moving the contour towards regions where $D$ is low.

\subsubsection{Internal terms}
\label{sec:Eint}

In the literature, the values of $\alpha$ and $\beta$ are generally a single scalar, meaning that the penalization has the same strength in all parts of the object. This leads to a trade-off between over-smoothing corner regions and under-smoothing others.
We avoid this trade-off by assigning different $\beta$ penalizations to each pixel, depending on which part of the object lies underneath.

The internal energy $E_{int} = \alpha\big(\mathbf{x},(\mathbf{y}_s)\big) |\mathbf{y}'|^2  + 	\beta\big(\mathbf{x},(\mathbf{y}_s)\big)|\mathbf{y}''|^2 $ penalizes the length (membrane term) and curvature (thin plate term) of the polygon. In order to obtain the direction of steepest descent, we can express the internal energy as a function of finite differences:
\begin{equation}
\small 
E_{int} 
=
\sum_{s=0}^L
\alpha(\mathbf{y}_{s})\big|\frac{\mathbf{y}_{s+1} - \mathbf{y}_{s}}{\Delta s}\big|^2
+
\beta(\mathbf{y}_{s})\big|\frac{\mathbf{y}_{s+1} - 2\mathbf{y}_{s}+\mathbf{y}_{s-1}}{\Delta s^2}\big|^2,
\end{equation}
and compute the derivative of $E_{int}$ w.r.t. the coordinates of node $s$, $\mathbf{y}_s$, expressed as a sum of scalar products:
\begin{multline}
\label{eq:dedy}
\frac{\partial E_{int}}{\partial \mathbf{y}_{s}} 
=
\frac{2}{\Delta s}[-\alpha_{s-1},\alpha_{s-1}+\alpha_{s},-\alpha_{s}]\cdot[\mathbf{y}_{s-1},\mathbf{y}_{s},\mathbf{y}_{s+1}]^\top
\\+
\frac{2}{\Delta s^2}
[\beta_{s-1},-2\beta_{s}-2\beta_{s-1},\beta_{s-1}+4\beta_{s}+\beta_{s+1},\\
-2\beta_{s+1}-2\beta_{s},\beta_{s+1}]\cdot
[\mathbf{y}_{s-2},\mathbf{y}_{s-1},\mathbf{y}_{s},\mathbf{y}_{s+1},\mathbf{y}_{s+2}]^\top.
\end{multline}
The Jacobian matrix (in this case with two column vectors) can then be expressed as a matrix multiplication:

\begin{equation}
\frac{\partial E_{int}}{\partial \mathbf{y}} 
=
(A+B)\mathbf{y}
\end{equation}

where $A(\alpha)$ is a tri-diagonal matrix and $B(\beta)$ is a penta-diagonal matrix.

\subsubsection{Balloon term}
\label{sec:balloon}

The original balloon term~\cite{cohen1991active} consists of adding an outwards force of constant magnitude in the normal direction of each node, thus inflating the contour. 
As with the $\beta$ term, we propose to increase its flexibility by allowing it to take a different value at each image location. 

In~\cite{cohen1991active}, the balloon term is only considered as a force added after the direction of steepest descent for the other energy terms has been computed. In DSAC, the SSVM formulation requires to express it in the form an energy.

The normal direction to the contour at $\mathbf{y}_s$ follows the vector:
\begin{equation}
\mathbf{n}_{s} = \big[ \mathbf{y}_{s+1}-\mathbf{y}_{s-1}\big]_{+90^o} = \big[v_{n+1}-v_{n-1},u_{n-1}-u_{n+1}\big].
\end{equation}
This can be rewritten such that the whole set of $L$ normal vectors is expressed as:
\begin{equation}
\mathbf{n} = \Big[C\mathbf{v}, \mathbf{u}^\top C\Big]
\end{equation}
where $C$ is a tri-diagonal matrix with $0$ in the main diagonal, $1$ in the upper diagonal and $-1$ in the lower diagonal.

Integrating this expression with respect to $\mathbf{u}$ and $\mathbf{v}$, we obtain the scalar $E_b$, corresponding to the polygon's area (by the shoelace formula to compute the area of a polygon):

\begin{equation}
E_{b} = \mathbf{u}^\top C \mathbf{v} = \int\int_{u,v\in \Omega(\mathbf{y})} du dv
\end{equation}

Instead of maximizing the area of the polygon, which would be the result of pushing nodes in the normal direction, we propose to use a more flexible term that maximizes the integral of the values of a map $\kappa(\mathbf{x})\in\mathbb{R}^{M\times N}$ over the area enclosed by the contour, $\Omega(\mathbf{y})$. If we discretize the integral to the pixel values that conform $\kappa$, we obtain:

\begin{equation}
E_{k} = \sum_{u,v\in \Omega(\mathbf{y})} \kappa(u,v)
\end{equation}

After this modification we need to recompute the force form of this term by finding the $L\times 2$ Jacobian matrix $[\frac{\partial E_k}{\partial u_s},\frac{\partial E_k}{\partial v_s}],$ $s \in [1,L]$.

\begin{figure}[h]
\centering
\begin{tabular}{cc}
\includegraphics[width=30mm]{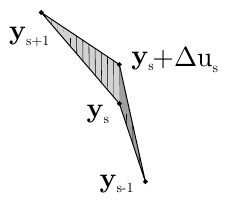}
&
\includegraphics[width=30mm]{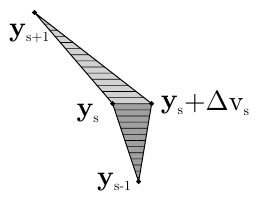}\\
a) & b)\\
\end{tabular}
\caption{A perturbation of $\mathbf{y}_s$ in either the $u$ or $v$ direction would result in a change in area highlighted as two shaded triangles sharing the same base.}
\label{fig:area}
\end{figure}

This corresponds to how a perturbation in $u_s$ and $v_s$ would affect $E_k$. Since the perturbations are considered to be very small, we assume that the distribution of the $\kappa(u,v)$ values along the segments $[\mathbf{y}_s,\mathbf{y}_{s+1}]$ and $[\mathbf{y}_{s-1},\mathbf{y}_{s}]$ will be identical to the one in $[\mathbf{y}_s+\Delta\mathbf{y},\mathbf{y}_{s+1}]$ and $[\mathbf{y}_{s-1},\mathbf{y}_{s}+\Delta\mathbf{y}]$, respectively. As shown in Fig.~\ref{fig:area}, this boils down to summing a series of trapezoid areas, forming the two depicted triangles, each one weighted by its assigned $\kappa$ value.

In Fig.~\ref{fig:area}a, both triangles have bases of length $\Delta u_s$ and heights $v_{s-1}-v_s$ and $v_{s+1}-v_s$, while in Fig.~\ref{fig:area}b the bases are $\Delta v_s$ and the heights $u_{s-1}-u_s$ and $u_{s+1}-u_s$.

To obtain the $\kappa$ weighted areas in Fig.~\ref{fig:area}a, we compute:
\begin{multline}
\Delta E_k
=
\frac{\Delta u_s}{v_{s-1}-v_s}\int_{h=0}^{v_{s-1}-v_s}
h \kappa(h) dh 
 +\\
\frac{\Delta u_s}{v_{s+1}-v_s}\int_{h=0}^{v_{s+1}-v_s}
h \kappa(h) dh,
\end{multline}
and therefore the force term we need for inference is:
\begin{multline}
\frac{\partial E_k}{\partial u_s}
=
\frac{1}{v_{s-1}-v_s}\int_{h=0}^{v_{s-1}-v_s}
h \kappa(h) dh
+\\
\frac{1}{v_{s+1}-v_s}\int_{h=0}^{v_{s+1}-v_s}
h \kappa(h) dh
\end{multline}
The same for Fig.~\ref{fig:area}b can be obtained by swapping $u$ and $v$.

These derivatives point in the normal direction when the values of $\kappa$ are equal in all locations.

\subsection{Active contour inference and implementation}
\label{sec:inference}

When solving the active contour inference, Eq.~(\ref{eq:snakes}), the four energy terms can be split into external terms $E_{ext}$: the data ($D$) and balloon energies ($E_k$); and internal terms $E_{int}$: the energies penalizing length ($\alpha$) and curvature ($\beta$). Since $E_{int}$ depends only on the contour $\mathbf{y}$, we can find an update rule that minimizes it on the new time time step:

\begin{equation}
\mathbf{y}^{t+1} = 
\mathbf{y}^{t} - \frac{d E_{ext}}{d \mathbf{y}^{t}} - (A+B)\mathbf{y}^{t+1}.
\end{equation}

If we solve this expression for $\mathbf{y}^{t+1}$, we obtain:

\begin{equation}
\mathbf{y}^{t+1} = (I+A+B)^{-1}\Big(\mathbf{y}^{t} - \frac{d E_{ext}}{d \mathbf{y}^{t}}\Big).
\end{equation}

With $I$ being the identity matrix. An efficient implementation of the ACM inference is critical for the usability of the method, since thousands of iterations are typically required by CNNs to be trained, and the ACM inference has to be performed at each iteration. We have implemented the described locally penalized ACM using a Tensorflow graph. The typical inference time is under 50 ms on a single CPU for the settings used in this paper. 

\subsection{Structured SVM loss}
\label{sec:loss}

Since no ground truth is available for the penalization terms, we frame the problem as structured prediction, in which loss augmented inference is used to generate negative examples to complement the positive examples of the ground truth polygons. The weights of the energy terms can then be modified such that the energy corresponding to the ground truth is lowered, while the one of the loss augmented results, which are presumed to be wrong, is increased.


Given a collection of ground truth pairs $(\mathbf{y^i},\mathbf{x^i}) \in \mathcal{Y}\times\mathcal{X}, i=1\dots N$, and a task loss function $\Delta(\mathbf{y},\mathbf{\hat{y}})$,
we would like to find the CNN parameters $\omega$ such that, by optimizing Eq.~(\ref{eq:snakes}) and thus obtaining the inference result:
\begin{equation}
\mathbf{\hat{y}}^i=\arg\min_{\mathbf{y}\in\mathcal{Y}}E(\mathbf{y},\mathbf{x},\omega)
\end{equation}
one could expect a small $\Delta(\mathbf{y}^i,\mathbf{\hat{y}}^i)$. The problem becomes:

\begin{equation}
\hat{\omega} = 
\arg\min_\omega\sum_i \Delta(\mathbf{y}^i,\arg\min_{\mathbf{y}\in\mathcal{Y}}E(\mathbf{y},\mathbf{x},\omega))
\end{equation}

Since $\Delta(\mathbf{y}^i,\mathbf{\hat{y}}^i)$ could be a discontinuous function, we can substitute it by a continuous and convex upper bound, such as the hinge loss.
By adding an $\ell_2$ regularization and summing for all training samples, this becomes the max-margin formulation:

\begin{align}
\label{eq:ssvm_loss2}
&\mathcal{L}(\mathcal{Y},\mathcal{X},\omega) = 
\frac{1}{2} \|\omega\|^2 + \\
&C\sum_i \bigg(\max_{\mathbf{y}\in\mathcal{Y}}\big[0,\Delta(\mathbf{y},\mathbf{y}^i) - E(\mathbf{y},\mathbf{x}^i;\omega) + E(\mathbf{y}^i,\mathbf{x}^i;\omega)\big]\bigg).\nonumber
\end{align}

Since $\mathcal{L}(\mathcal{Y},\mathcal{X},\omega)$ is convex but not differentiable, we compute the subgradient, which requires to find the most penalized constraint with the current $\omega$:

\begin{equation}
\hat{\mathbf{y}}^i = \arg\max_{\mathbf{y}\in\mathcal{Y}}\big[\Delta(\mathbf{y},\mathbf{y}^i) - E(\mathbf{y},\mathbf{x}^i;\omega)\big]
\label{eq:loss_inference}
\end{equation}

This means to first run the ACM using the current $\omega$ and an extra term corresponding to the loss $\Delta(\mathbf{y},\mathbf{y}^i)$. Once we obtain $\hat{\mathbf{y}}^i$, we can then compute the subgradient as:

\begin{equation}
\frac{\partial\mathcal{L}(\mathcal{Y},\mathcal{X},\omega)}{\partial \omega} = 
\omega + C \sum_i \big( 
\frac{\partial E(\mathbf{y}^i,\mathbf{x}^i;\omega)}{\partial\omega} - 
\frac{\partial E(\hat{\mathbf{y}}^i,\mathbf{x}^i;\omega)}{\partial\omega}
 \big)
\end{equation}

We compute the subgradients of the loss with respect to each of the four outputs as

\begin{align}
\label{eq:grad_D}
&\frac{\partial\mathcal{L}(\mathbf{y}^i,\mathbf{x}^i,\omega)}{\partial D_{\omega}(\mathbf{x}^i)}
=
[(u,v)\in\mathbf{y}^i]
-
[(u,v)\in\hat{\mathbf{y}}^i]
\end{align}

\begin{align}
\label{eq:grad_A}
&\frac{\partial\mathcal{L}(\mathbf{y}^i,\mathbf{x}^i,\omega)}{\partial \alpha_{\omega}(\mathbf{x}^i)}
= \\\
&\Big|\frac{\partial\mathbf{y}^i(u,v)}{\partial s}\Big|^2
[(u,v)\in\mathbf{y}^i] 
-
\Big|\frac{\partial\hat{\mathbf{y}}^i(u,v)}{\partial s}\Big|^2
[(u,v)\in\hat{\mathbf{y}}^i] \nonumber
\end{align}

\begin{align}
\label{eq:grad_B}
&\frac{\partial\mathcal{L}(\mathbf{y}^i,\mathbf{x}^i,\omega)}{\partial \beta_{\omega}(\mathbf{x}^i)}
= \\ 
&\Big|\frac{\partial^2\mathbf{y}^i(u,v)}{\partial s^2}\Big|^2
[(u,v)\in\mathbf{y}^i]
-
\Big|\frac{\partial^2\hat{\mathbf{y}}^i(u,v)}{\partial s^2}\Big|^2
[(u,v)\in\hat{\mathbf{y}}^i]  \nonumber
\end{align}

\begin{equation}
\label{eq:grad_K}
\frac{\partial\mathcal{L}(\mathbf{y}^i,\mathbf{x}^i,\omega)}{\partial \kappa_{\omega}(\mathbf{x}^i)}
=
[(u,v)\in\Omega(\mathbf{y}^i)]
-
[(u,v)\in\Omega(\hat{\mathbf{y}}^i)].
\end{equation}

In the above equations,  $[\cdot]$ represents the Iverson bracket. Finally, we can get $\frac{\partial\mathcal{L}(\mathcal{Y},\mathcal{X},\omega)}{\partial \omega}$ using the chain rule and modifying each CNN parameter $\omega$ applying:

\begin{equation}
\omega_{t+1} = \omega_{t} - \eta \frac{\partial\mathcal{L}(\mathcal{Y},\mathcal{X},\omega)}{\partial \omega},
\end{equation}

which will simultaneously decrease $E(\mathbf{y}^i,\mathbf{x}^i;\omega)$ and increase $E(\hat{\mathbf{y}}^i,\mathbf{x}^i;\omega)$, thus making a better solution more likely when performing inference anew.

\paragraph{Task loss}

\begin{figure}[h!]
\centering
\includegraphics[width=0.5\linewidth]{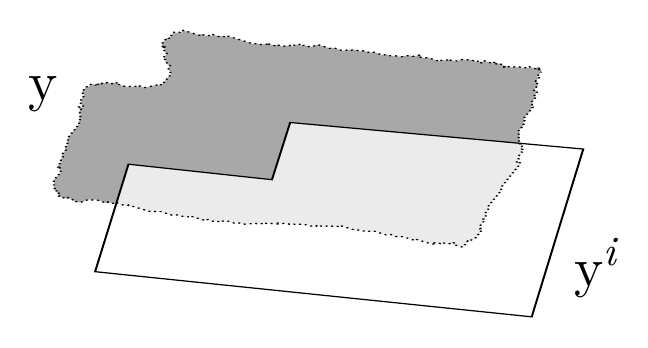}
\caption{When training we encourage a high task loss (IoU) by modifying the balloon term $E_\kappa$, adding a negative constant to $\kappa$ at the nodes of the prediction $\mathbf{y}$ inside the ground truth $\mathbf{y^i}$ (light gray), and a positive constant to those outside (dark gray).}
\label{fig:taskLoss_iou}
\end{figure}

The task loss $\Delta(\mathbf{y},\mathbf{y}^i)$ defines the actual objective we want to solve with the SSVM loss. Since it's the most common metric in instance segmentation, we employ the Intersection-over-Union (IoU) between the prediction $\mathbf{y}$ and the ground truth $\mathbf{y}^i$. Note that optimizing for IoU can be split into maximizing the intersection while minimizing the union. During training, this allows us to simply add a negative value during training to the $\kappa$ map at the locations within the ground truth and a positive outside to obtain a loss-augmented inference (see Fig. \ref{fig:taskLoss_iou}).

\section{Experiments}
\label{sec:experiments}

We test the proposed DSAC method for building footprint extraction from overhead images. We consider two settings: \emph{manual initialization}, where the user provides a single click near the center of the building and \emph{automatic initialization}, where an instance segmentation algorithm is used to generate the initial polygons. The first setting is tested in two datasets, \emph{Vaihingen} and \emph{Bing Huts}, while the second is tested in the \emph{TorontoCity} dataset~\cite{wang2016torontocity}. The three datasets are detailed in the respective sections.

\subsection{CNN architecture and general setup}
To learn the ACM energy terms, we use a CNN architecture similar to the Hypercolumn model in~\cite{hariharan2015hypercolumns}. The input consists of a patch cropped around each initialization polygon and resized an image of fixed size for each dataset. The first layer consists of $7\times 7$ convolutions, the second of $5\times 5$ and all subsequent layers are of size $3\times 3$. All the convolutional layers are followed by ReLu, batch normalization and $2\times 2$ max-pooling. The number of filters is increased with the depth: $32$, $64$, $128$ ,$128$, $256$ and $256$ for the six blocks. The output tensors of all the layers are then upsampled to the output size and concatenated. After this, a two-layer MLP with 256 and 64 hidden units is used to predict the four output maps: $D(\mathbf{x})$, $\alpha(\mathbf{x})$, $\beta(\mathbf{x})$ and $\kappa(\mathbf{x})$. We use this architecture for all datasets, with the exception of the \emph{Bing huts} dataset, for which we skip the last two convolutional layers. In all cases, we use the Adam optimizer with a learning rate of $10^{-4}$. We augment the data with random rotations. 
The number of ACM iterations is set to 50 in all the experiments, and the number of nodes is set to $L=60$ in \emph{Vaihingen} and \emph{TorontoCity} and $L=20$ in \emph{Bing huts}.

\subsection{Manual initialization}

In this setting, the detection step is done manually by visual inspection. The only input required from the user is a single click to indicate the approximate center of the building. Two datasets are considered:

\paragraph{Vaihingen buildings}
The dataset consists of 168 buildings extracted from the training set of the ISPRS ``2D semantic labeling contest''\footnote{\url{http://www2.isprs.org/commissions/comm3/wg4/semantic-labeling.html}}. The images have three bands, corresponding to near infrared, red and green wavelengths, and a resolution of 9 cm. We used $100$ buildings to train the models and the remaining $68$ as a test set.

\begin{figure}[!h]
\begin{tabular}{cc}
\includegraphics[height=0.43\linewidth]{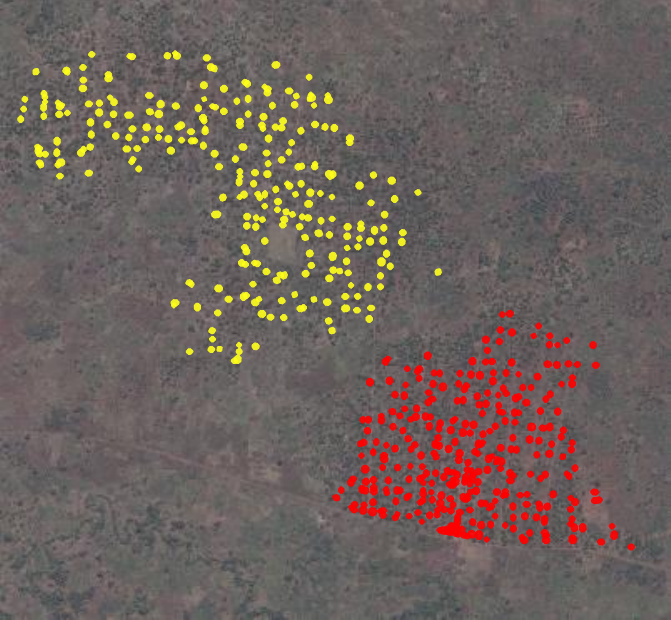} &
\includegraphics[height=0.43\linewidth]{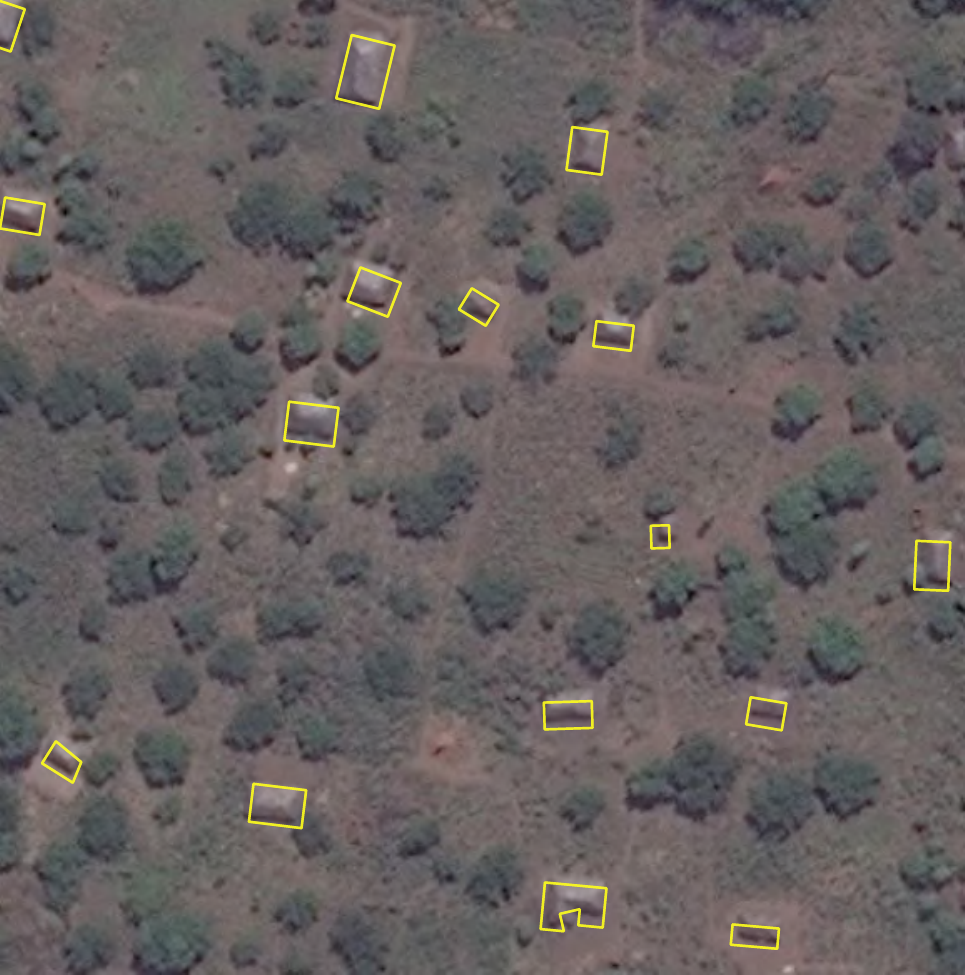}
\end{tabular}
\caption{Left: Overview of the 4 km$^2$ area covered by the Bing huts dataset. The training instances are higlighted in red and the test ones in yellow. Right: detail of the test set.}
\label{fig:bing_over}
\end{figure}

\begin{figure*}[ht]
\centering
\includegraphics[width=\linewidth]{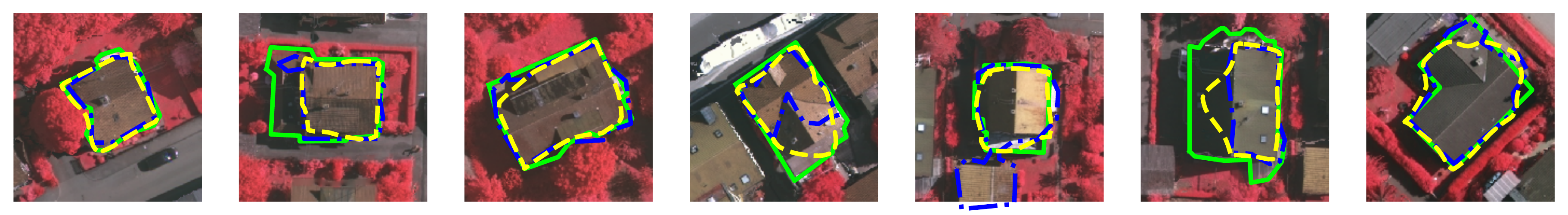}
\includegraphics[width=\linewidth]{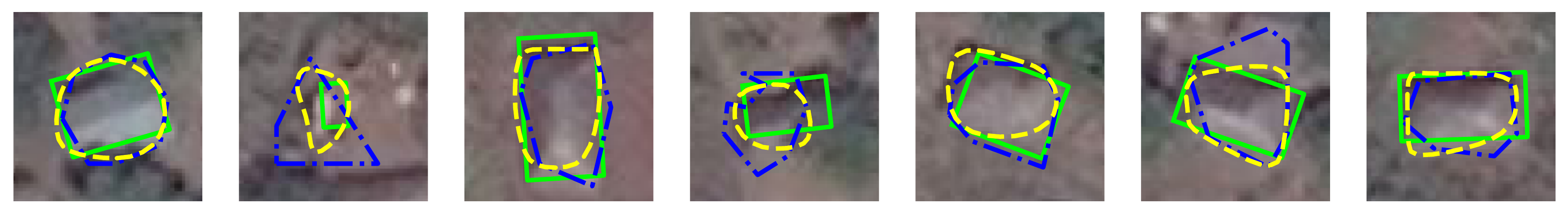}
\includegraphics[width=0.984\linewidth]{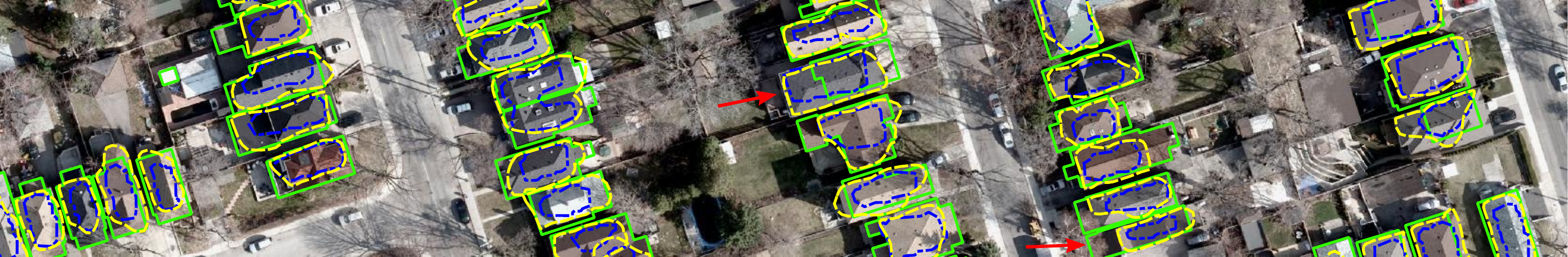}
\caption{Examples of test set buildings in the Vaihingen (top row), Bing huts (middle row) and TorontoCity (bottom row) datasets. \textcolor{green}{Ground truth} in solid green line, \textcolor{blue}{baseline} result in dash-dot blue and our \textcolor{yellow}{active contour} result in dashed yellow. Note that some of the ground truth polygons in the TorontoCity dataset are shifted (red arrows).}
\label{fig:results}
\end{figure*}

\paragraph{Bing huts}
The dataset consists of $605$ individual huts visible on Bing maps aerial imagery at a resolution of $30$ cm, over a rural area in Tanzania. See Fig.~\ref{fig:bing_over} for an overview of the study area and Fig.~\ref{fig:results} for a full resolution subset. The ground truth building footprints have been obtained from OpenStreetMap\footnote{\url{http://www.openstreetmap.org}}. A total of $335$ images of size $80\times 80$ pixels are used to train the models and the remaining $270$ to test. The lower spatial resolution, low contrast between the buildings and the surrounding soil, as well as the high level of label noise make \emph{Bing huts} a very challenging dataset.


We compare \devis{the results obtained by}{} DSAC \devis{on the two datasets}{} against a baseline where we train a CNN with the same architecture used by DSAC, but with a 3-class cross entropy loss with classes: building, building boundary, background. The boundary class is added to help the model focus on learning the shapes of the buildings. In this case, the click from the user is used to select the nearest connected region that has been labeled as building and treat it as the instance prediction.

\subsection{Automatic initialization}
Although the manual initialization only requires a single click from the user, it can still be a tedious task for large scale datasets. Existing instance segmentation algorithms, such as the recently proposed Deep Watershed Transform (DWT)~\cite{Bai17}, can be used instead to initialize the active contours. These methods have a good recall, but tend to undersegment the objects and to lose detail near to the boundaries. To compensate for this effect, the authors of~\cite{Bai17} apply a morphology-based post-processing step. We test the possibility of initializing the ACM within DSAC with the results obtained by~\cite{Bai17} on the TorontoCity building instance segmentation dataset~\cite{wang2016torontocity}, with around $28000$ instances for training and $12000$ for testing. The ACM contours are initialized with the output of the Deep Watershed Transform (DWT)~\cite{Bai17}, the current state-of-the-art in terms of IoU. Two initialization polygon types are considered: the raw DWT output and the post-processed versions used in~\cite{wang2016torontocity}. We also consider a third variant, where the raw DWT is used at train time and the post-processed one for inference at test time: this variant is based on the intuition that making the problem harder at train time, in addition to using the loss augmentation, helps learning a better energy function.

\begin{figure*}[h]
\includegraphics[width=17cm]{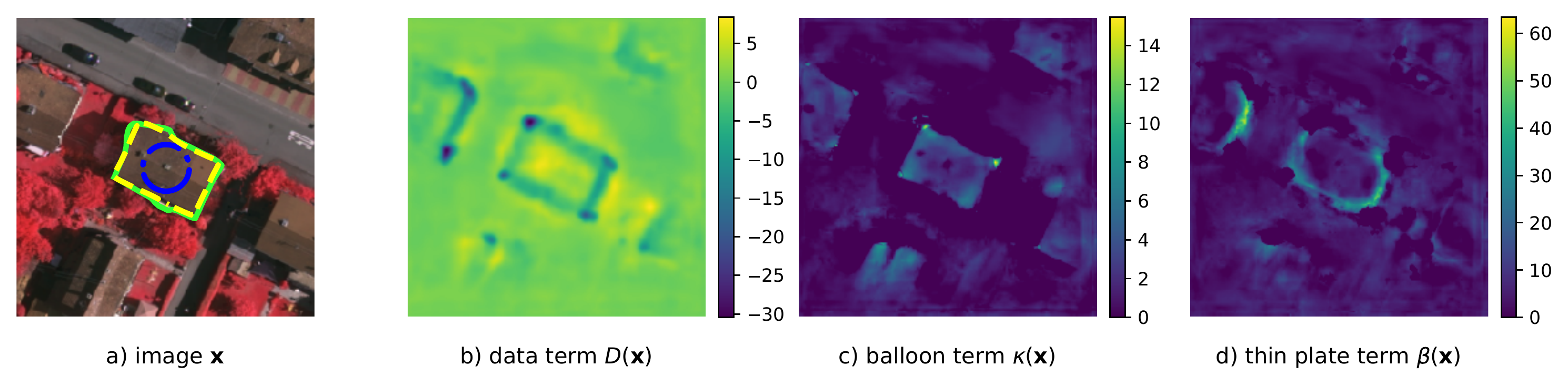}
\caption{a) Image from the Vaihingen test set. The initial contour is in blue and the result in yellow, with the ground truth in green. b) Data term $D(\mathbf{x})$, where we can observe regions of lower energy along the boundary of the building. c) The balloon term $\kappa(\mathbf{x})$ has learned to produce positive values only inside the building, especially next to corners. d) In the thin plate term $\beta(\mathbf{x})$, we see that the curvature tends to be less penalized close to the building's corners. The membrane term provided by the model in this example was $\alpha(\mathbf{x})=0.74$}
\label{fig:vai_output}
\end{figure*}

\section{Results and discussion}
\label{sec:resultsAndDiscussion}


\paragraph{Manual initialization} Table~\ref{tab:vai_bing} reports the average Intersection over Union (IoU) for the two datasets. Since the ground truth shift noise in the Bing huts dataset makes the IoU assessment untrustworthy, the root mean square error (RMSE in $m^2$) committed when estimating the area of the building footprints is also reported. DSAC significantly improves the baseline in terms of IoU for both datasets. This ablation study confirms the need to allow $\kappa$ and $\beta$ to vary locally (as opposed to having a single value for the whole image), while $\alpha$ can be treated as a single value without loss of performance. It also highlights the importance of the balloon term for the convergence of the contour.

\begin{table}[!h]
\centering
\begin{tabular}{c|c|c|c|}
\cline{2-4}
 & \multicolumn{2}{c|}{Average IoU} & RMSE \\ \cline{2-4} 
 & Vaihingen & Bing huts & Bing huts \\ \hline
\multicolumn{1}{|l|}{CNN Baseline} & 0.78 & 0.56 & 23.9 \\ \hline
\multicolumn{1}{|l|}{DSAC (ours)} & \textbf{0.84} & \textbf{0.65} & \textbf{13.4} \\ \hline
\multicolumn{1}{|l|}{DSAC (scalar $\kappa$, $\beta$)} & 0.64 & 0.60 & 19.1 \\ \hline
\multicolumn{1}{|l|}{DSAC (no $\kappa$)} & 0.63 & 0.42 & 31.2 \\ \hline
\multicolumn{1}{|l|}{DSAC (local $\alpha$)} & 0.83 & \textbf{0.65} & \textbf{13.4} \\ \hline
\end{tabular}
\caption{Results on the test set for the manual initialization experiments, reported as average intersection over union (IoU, left) and area estimation (\emph{Bing huts} only), with RMSE in $m^2$ (right).}
\label{tab:vai_bing}
\end{table}

Examples of segmentation results for the Vaihingen dataset (Fig.~\ref{fig:results}, top row) show that the learned priors do indeed promote smooth, straight edges while often allowing for sharp corners. By looking at the predicted energy terms in Fig.~\ref{fig:vai_output} we observe that the model focuses on the corners by producing very low $D$ values close to them, while predicting high $\kappa$ inside the building next to the corners and a sharp drop to $0$ on the outside. Moreover, the smoothness term $\beta$ is close to $0$ at the corners and high along the edges. 

In the Bing huts dataset results (Fig.~\ref{fig:results}, bottom row), the biggest jump in performance can be seen in the area estimation metric. DSAC still tends to oversmooth the shapes, probably since it is unable to learn the location of corners due to the ground truth shift noise inherent to OpenStreetMap data, but manages to converge to polygons of the correct size, most probably because it learns to balance the balloon ($\kappa$, promoting large areas) and the membrane ($\alpha$, promoting short contours) terms.


\paragraph{Automatic initialization} Table~\ref{tab:toronto} reports the results obtained on the TorontoCity dataset using two metrics: the IoU-based weighted coverage (``WeighCov'') and the shape similarity PolySim~\cite{wang2016torontocity}.  
Besides DWT, we also compare DSAC against the results of building footprint segmentation with FCN and ResNet, as reported in~\cite{wang2016torontocity}. We observe an improvement with respect to DWT of both metrics. DSAC obtains the best weighted coverage scores irrespectively of the initialization strategy. Interestingly, the best results are obtained by the hybrid initialization using raw DWT at training time and post-processed DWT polygons at test  time.
This suggests that our intuition about making the model work harder at train time is correct and seems to complement the use of a task loss in the SSVM loss.
Finally, segmentation examples are shown in the last row of Fig.~\ref{fig:results}: DSAC (in yellow) consistently returns a more desirable segmentation with respect to DWT (in blue), closer to the ground truth polygon (in green). 
Although we can still see oversmoothing in our results, note how an important amount of shift noise is also present in some instances, making the DSAC result more plausible than the ground truth in a few cases (red arrows).

\begin{table}[h]
\centering
\begin{tabular}{c|c|c|}
\cline{2-3}
 & WeighCov & PolySim \\  \hline
 \multicolumn{1}{|l|}{FCN~\cite{long2015fully}} & 0.46 & \textbf{0.32} \\ \hline
 \multicolumn{1}{|l|}{ResNet~\cite{he2016deep}} & 0.40 & 0.29 \\ \hline\hline
\multicolumn{1}{|l|}{DWT, raw~\cite{Bai17} (\texttt{RW})} & 0.42 & 0.20 \\ \hline
\multicolumn{1}{|l|}{DWT, postproc. (\texttt{PP})} & 0.52 & 0.24 \\ \hline\hline
\multicolumn{1}{|l|}{DSAC (init.: train \texttt{RW} / test \texttt{RW})} & 0.55 & 0.26 \\ \hline
\multicolumn{1}{|l|}{DSAC (init.: train \texttt{PP} / test \texttt{PP})} & 0.57 & 0.26 \\ \hline
\multicolumn{1}{|l|}{DSAC (init.: train \texttt{RW} / test \texttt{PP})} & \textbf{0.58} & 0.27 \\ \hline
\end{tabular}
\caption{Results of the proposed DSAC and the methods reported in~\cite{wang2016torontocity} on the validation set of the TorontoCity dataset, containing over $12000$ detected building instances. Two ACM initializations, \texttt{RW} (\cite{Bai17}) and \texttt{PP} (\cite{Bai17} post-processed), are compared.}
\label{tab:toronto}
\vspace{-0.5cm}
\end{table}

\section{Conclusion}
\label{sec:conclusion}

We have shown the potential of embedding high-level geometric processes into a deep learning framework for the segmentation of object instances with strong shape priors, such as buildings in overhead images.
The proposed Deep Structured Active Contours (DSAC) uses a CNN to predict the energy function parameters for an Active Contour Model (ACM) such as to make its output close to a ground truth set of polygonal footprints. The model is trained end-to-end by bringing the ACM inference into the CNN training schedule and using the ACM's output and the ground truth polygon to assess a structured loss that can be used to update the CNN's parameters using back-propagation.
DSAC opens up the possibility of using a large collection of energy terms encoding for different priors, since an adequate balance between them is learned automatically. The main limitation of our model is that the initialization is assumed to be given by some external method and is therefore not included in the learning process.

Results in three different datasets, which include a $10\%$ relative improvement over the state-of-the-art on the \emph{TorontoCity} dataset, show that combining the bottom-up feature extraction capabilities of CNNs with the high-level constraints provided by ACMs is a promising path for instance segmentation when strong geometric priors exist.

\newpage 

{\small
\bibliographystyle{ieee}
\bibliography{bib}
}

\end{document}